\documentclass[a4paper,conference]{IEEEtran}
\IEEEoverridecommandlockouts
\usepackage{cite}
\usepackage{amsmath,amssymb,amsfonts}
\usepackage{algorithmic}
\usepackage{graphicx}
\usepackage{textcomp}
\usepackage{xcolor}
\usepackage{hyperref}
\def\BibTeX{{\rm B\kern-.05em{\sc i\kern-.025em b}\kern-.08em
    T\kern-.1667em\lower.7ex\hbox{E}\kern-.125emX}}

\IEEEpubid{%
  \parbox{\textwidth}{%
    \vspace{5mm}
    \fontsize{8.5}{10}\selectfont\raggedright
    This paper is accepted by the 2026 IEEE International Conference on
    Computer Vision and Machine Intelligence (CVMI).
    Copyright \copyright{}2026 IEEE. Personal use of this material is permitted.
    However, permission to use this material for any other purposes must be
    obtained from IEEE by sending an email to pubs-permissions@ieee.org.%
  }%
}

\begin{document}

\title{CVBL-CCTV-Face Dataset for Real-World Face Enhancement and Recognition in Surveillance}

\author{
\IEEEauthorblockN{Manish Das, Ajay Kumar Yadav, Shiv Ram Dubey, Satish Kumar Singh}
\IEEEauthorblockA{
Computer Vision and Biometrics Lab, Indian Institute of Information Technology, Allahabad\\
\textit{\{pmm2023001, rsi2022502, srdubey, sk.singh\}@iiita.ac.in}
}

}

\maketitle

\begin{abstract}
In real-world surveillance settings, facial captures from closed-circuit television systems suffer from degradations such as low resolution, motion blur, noise, and occlusions, which undermine dependable face recognition. In this study, we propose a framework for super-resolution and identity recognition, aimed at surveillance applications. For this purpose, we present \textbf{the CVBL-CCTV-Face Dataset} which is a dataset of real-world images taken in an unpaired manner where each pair consists of a low-quality (LQ) image of a face captured by a 8MP IP Dome CCTV camera together with a high-quality (HQ) reference image that is taken in controlled settings. In order to cope with the various problems posed by such datasets, we construct a modular system that involves face detection, deep learning-based super-resolution of low-quality face images, and identity recognition through L2-normalized embeddings extracted by an embedding model. Enhancement models are trained and tested according to three different setups, and the system is deployed for identity verification in a real-time unconstrained environment.
\end{abstract}

\begin{IEEEkeywords}
CCTV surveillance, image super-resolution, deep learning, real-time enhancement, unpaired dataset
\end{IEEEkeywords}
\section{Introduction}
\label{intro}
 
\begin{figure}[t]
    \centering
    \includegraphics[width=0.48\textwidth]{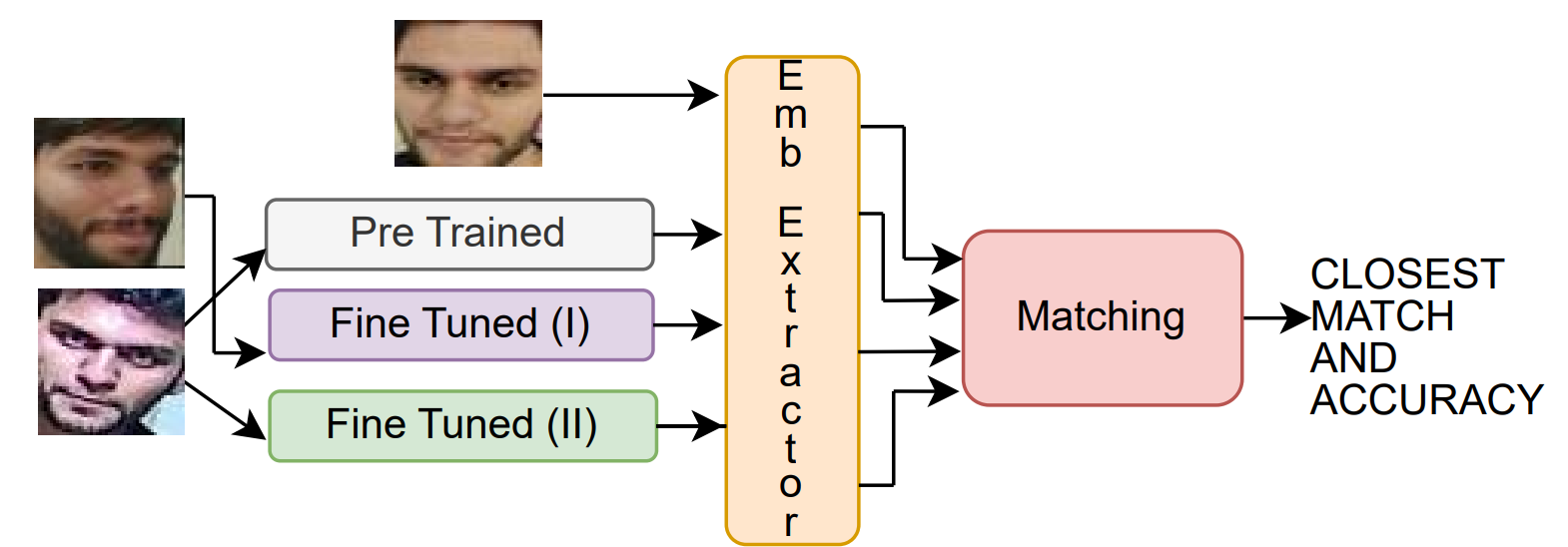}
    \caption{Training and evaluation configurations: (i) Pretrained models,
    (ii) Fine-tuning on synthetic HQ degradation, (iii) Fine-tuning on real
    LQ samples. \textit{Emb Extractor} denotes the embedding extraction
    model.}
    \label{fig:finetuning}
\end{figure}

\begin{figure}[t]
    \centering
    \includegraphics[width=0.48\textwidth]{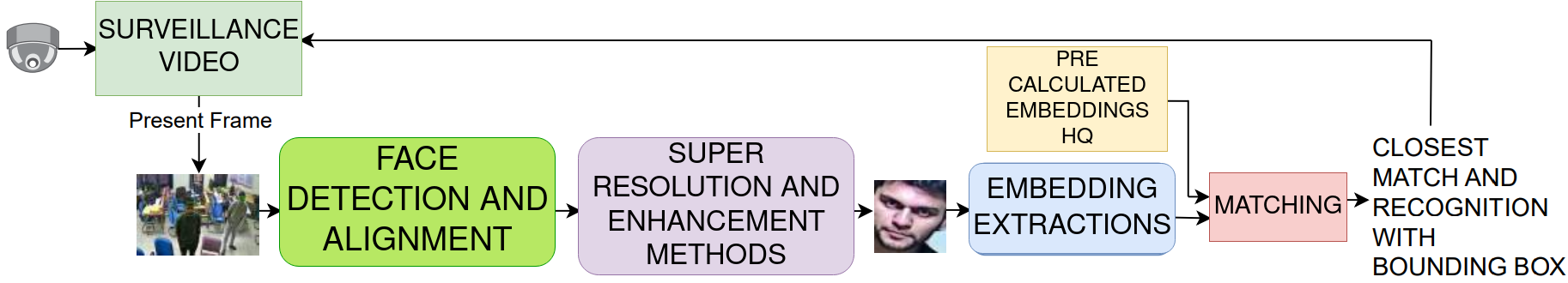}
    \caption{Real-time surveillance system with live CCTV stream processing,
    on-the-fly enhancement, and identity recognition.}
    \label{fig:realtime}
\end{figure}

Benchmarking systems for face recognition has largely been done on face images obtained from studios and synthetically degraded images. However, there has never been a benchmarking database for closed circuit television (CCTV) images containing faces with multiple uncontrolled degradations such as low resolution because of the distance between the object and the camera, motion blur due to motion of the subject, sensor noise, partial occlusion and wide pose variation. Thus, a need for a dataset with real LQ CCTV face images alongside HQ face images taken separately from the subjects in an unpaired manner arises. This motivates the creation of the \textbf{CVBL-CCTV-Face Dataset}. In this regard, we propose the \textbf{CVBL-CCTV-Face Dataset} which is a real world and unpaired collection of Low Quality (LQ) face images taken from a 8MP IP dome CCTV camera alongside High Quality (HQ) face images taken separately. The dataset includes two versions, \textbf{Version A}and \textbf{Version B,} differing in the number of HQ references per identity, enabling analysis under varying reference richness. To validate the dataset, we build a modular benchmark pipeline (see Fig. \ref{fig:simple_pipeline}) consisting of MTCNN \cite{mtcnn} face detection, transformer/diffusion-based enhancement, and L2-distance identity matching. Enhancement models are evaluated under three configurations as illustrated in Fig. \ref{fig:finetuning}: $(i)$ pretrained models, $(ii)$ fine-tuning on synthetically degraded HQ images, and $(iii)$ fine-tuning on real LQ samples. Finally, we demonstrate the  real-time surveillance system that processes live CCTV streams, enhancing detected faces and verifying identities with bounding box overlays (see
Fig. \ref{fig:realtime}). The key contributions of this work are:

\begin{figure}[t]
    \centering
    \includegraphics[width=\columnwidth]{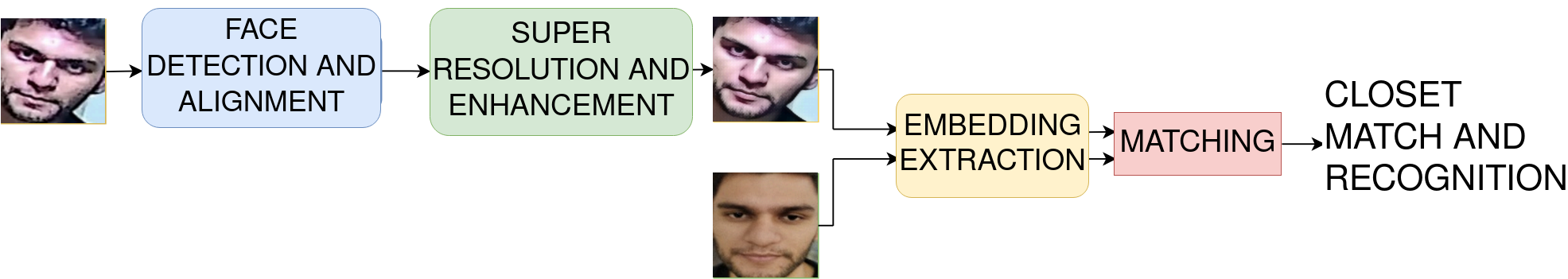}
    \caption{Overview of the proposed face enhancement and recognition
    pipeline integrating face detection, enhancement, and identity
    recognition.}
    \label{fig:simple_pipeline}
\end{figure}
 
\begin{itemize}
    \item \textbf{CVBL-CCTV-Face Dataset}: A real-world, unpaired dataset
    of 223 identities with authentic CCTV-degraded LQ images  and separately acquired HQ references. Unlike prior datasets that simulate degradation synthetically, ours captures compound real-world distortions with no spatial alignment between
    LQ–HQ pairs.
 
    \item \textbf{Benchmark Pipeline}: A validation framework with MTCNN
    face detection, four transformer/diffusion enhancement models, and L2
    embedding-based recognition, also deployed for real-time CCTV face enhancement and identification with overlaid bounding boxes and labels.
    \item \textbf{Training Configuration Analysis}: \IEEEpubidadjcol Three training setups
        (pretrained, synthetic HQ degradation fine-tuning, real LQ fine-tuning)
        evaluated on the dataset to characterize its challenges on surveillance performance.
\end{itemize}

\begin{figure}[!t]
    \centering

    \includegraphics[width=1.0\linewidth]{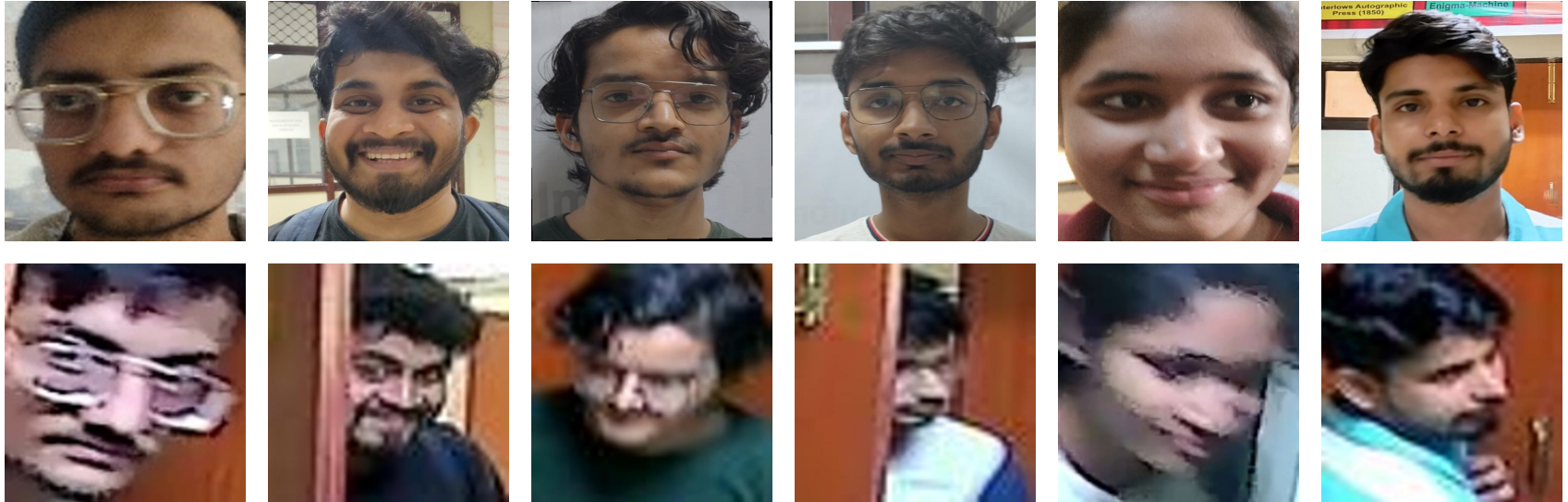}
    \caption{\textbf{ CVBL-CCTV-Face Dataset}: top row shows high-quality (HQ) references, bottom row shows corresponding low-quality (LQ) images.}
    \label{fig:samples}
\end{figure}

\section{Related Work}

Restoring faces from real-world surveillance captures is difficult because of complex degradations. Early super-resolution (SR) methods\cite{5,10,22,27} targeted pixel fidelity (PSNR, SSIM\cite{52}) by creating LQ–HQ pairs via bicubic downsampling and using architectures with dense/residual connections, recursion, non-local attention, and other advanced modules. However, these techniques struggle with the heterogeneous distortions found in authentic CCTV data. Further, methods like BSRGAN \cite{65} and RealESRGAN \cite{50} synthesize varied degradations, while perceptual \cite{17} and GAN-based losses \cite{24,48} improve visual quality at the cost of artifacts. Subsequent work mitigated these artifacts through frequency division \cite{25}, network optimization \cite{70}, adaptive loss re-weighting \cite{43,49}, and leveraging local statistics \cite{29}. Diffusion-based priors have recently advanced blind SR: StableSR \cite{46}, and DiffBIR \cite{30} adapt pretrained diffusion backbones. Recent architectures like PiSA-SR \cite{pisa} disentangle pixel-level and semantic enhancement using dual-LoRA branches, while TSD-SR \cite{tsd} aligns SR outputs with recognition utility via target score distillation. To improve efficiency, methods like SinSR \cite{51} and OSEDiff \cite{56} reduce iterative overheads while preserving output consistency. Additionally, specialized enhancements have been put forward in surveillance systems. Whereas the learning-based models such as  GFPGAN \cite{gfpgan} have proposed component-aware priors and VQFR \cite{vqfr} has used vector quantization to stabilize latent features, transformer models such as CodeFormer \cite{codefrmr} and RestoreFormer++ \cite{r++} make use of codebook priors and key–value memory for dynamic reconstructions.
But these models are mainly aimed at expression analysis and re-identification under relatively controlled environments, using synthetic degradations. All these studies underscore the importance of super-resolution in multimedia face applications. While there is progress in super-resolution and enhancements, the usage of synthetic degradations has limited the real-life application of these techniques. It therefore becomes necessary to develop data sets and techniques specifically designed for surveillance environments.

\begin{figure}[!t]
    \centering
    \includegraphics[width=\columnwidth, height=8cm]{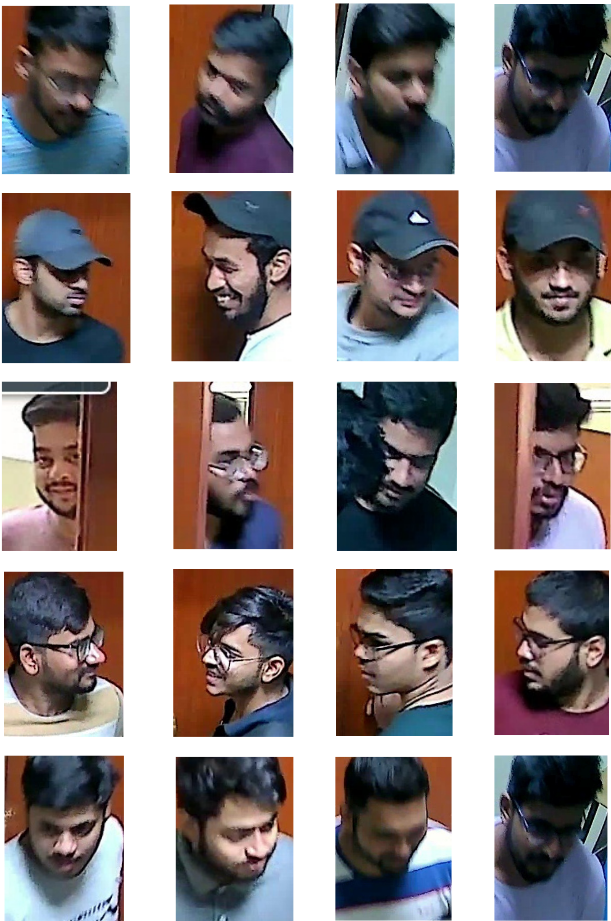}
    \caption{Examples of typical challenges in the \textbf{ CVBL-CCTV-Face Dataset}. Row 1: Blurred Faces due to motion and poor focus. Row 2: Faces with Caps causing occlusion. Row 3: Partial Faces obstructed by doors or other objects. Row 4: Side Faces with significant pose variation. Row 5: Head-Down Faces. }
    \label{fig:mtcn_challenges}
\end{figure}

\section{Proposed Dataset}
\label{sec:dataset}

\subsection{ CVBL-CCTV-Face Dataset}
\label{sec:cvbl_face_dataset}
\textbf{CVBL-CCTV-Face Dataset} is a collection of unpaired facial images in the real-world scenario of face enhancement and face recognition under surveillance degradations. An 8MP IP dome CCTV camera (model: EW-IP8050D) is employed to record the  low quality (LQ) images in unconstrained indoor environments, which may cause various artifacts, such as pose variation, illumination variation, partial occlusion and noise from the sensor. Considering the practical surveillance environment, the high quality (HQ) images are captured independently in a    controlled way. Some samples are demonstrated in Fig. \ref{fig:samples}. We provide the dataset in two different versions. \textbf{Version A} includes 102 identities, where each identity has one HQ image (totally 102 HQ images) and multiple LQ images (total 8090 images). The version represents the actual situation of having a limited number of HQ images. \textbf{Version B} includes 130 identities, where each identity has 15 to 25 HQ images (totally 2,612 HQ images) and 7,572 LQ images. The two versions share 9 identities and after deleting the duplication, we have a final count of 223 identities.  For experimentation, the final set is split with identity-level separation, where we have the test set, which includes 130 identities (1,581 LQ and 528 HQ images), and the remaining identities (14,081 LQ across 223 folders and 2,186 HQ images, augmented via standard transformations) make our training set.


\subsection{Challenges in the Dataset}
\label{sec:challenges_dataset}

\begin{figure}[!t]
    \centering
    \includegraphics[width=\columnwidth]{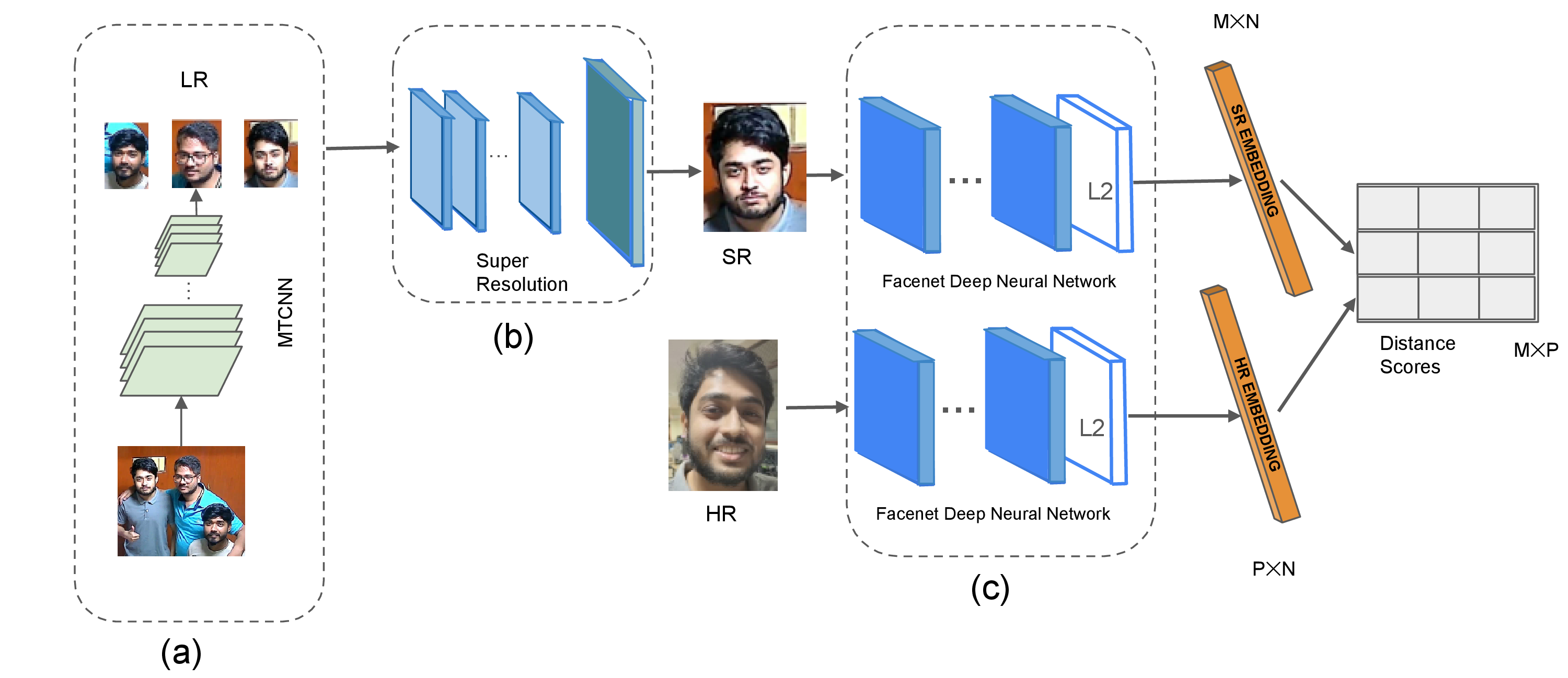}
    \caption{Proposed pipeline: (a) MTCNN face detection; (b) Super-resolution via facial enhancement models; (c) Inception-ResNet-v1 embedding extraction and L2 norm identity matching (M × 512 SR, P × 512 HR embeddings).}
    \label{fig:pipeline}
\end{figure}

\begin{figure}[t]
    \centering
    \includegraphics[width=\columnwidth]{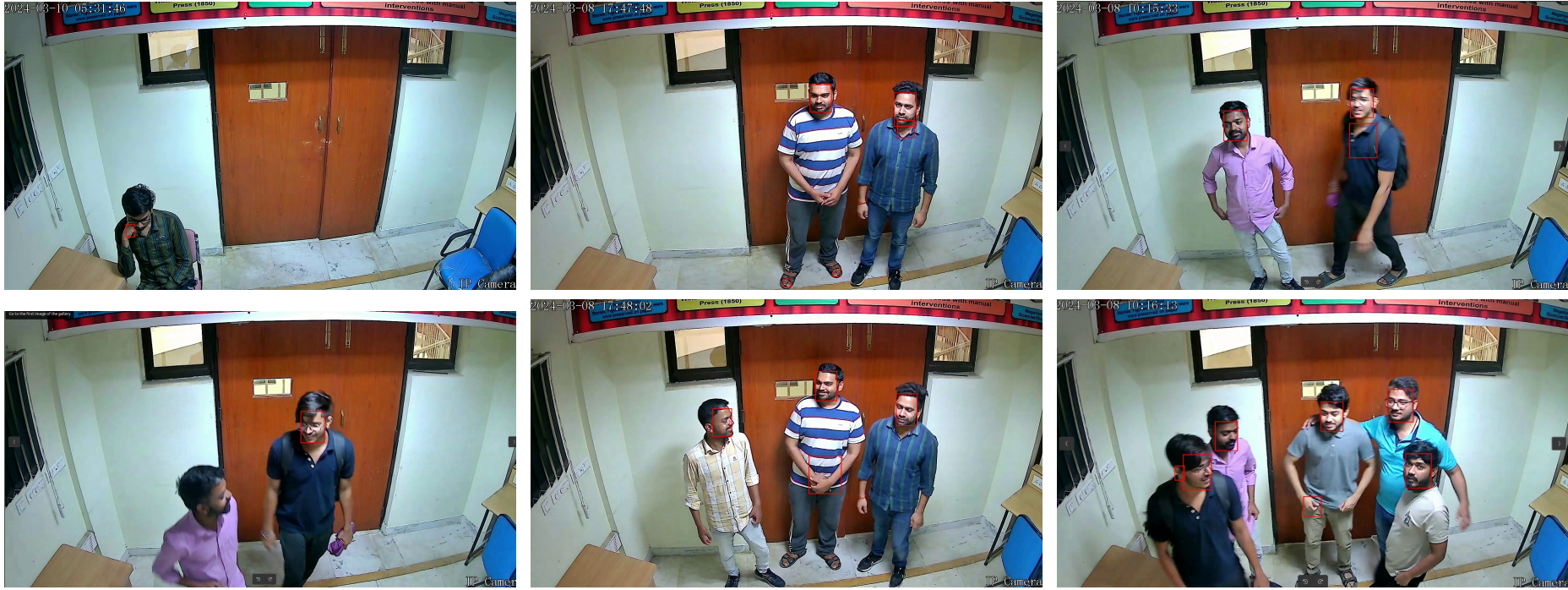}
    \caption{Illustration of various challenges encountered by MTCNN during face detection, highlighting the challenges in proposed \textit{CVBL-CCTV-Face Dataset}.}
    \label{fig:mtcnn_challenges}
\end{figure}

Real-world surveillance footage presents several inherent challenges (refer Fig. \ref{fig:mtcn_challenges}) for facial enhancement and recognition, as reflected in the \textbf{ CVBL-CCTV-Face Dataset} :

\begin{itemize}
    \item \textbf{Blurred Faces:} The facial clarity is often degraded by motion and defocus, which makes enhancement and recognition complicated.
    
    \item \textbf{Occlusions by Accessories:} Caps or hat items occlude many important landmarks, which affects identification.
    
    \item \textbf{Partially Visible Faces:} Walls or doors obstruct many facial regions, which affects restoration.
    
    \item \textbf{Pose Variations:} Nonfrontal views hinder facial structures and hamper models trained mainly on frontal faces.
    
    \item \textbf{Downward Head Tilts:} Key features like eyes and mouth become less visible when faces are angled downwards.
\end{itemize}

These challenges necessitate the development of robust enhancement models capable of handling diverse degradations and pose variations, ensuring reliable performance under unconstrained real-world surveillance conditions.

\section{Methodology}

\subsection{Face Enhancement and Recognition Pipeline}
\label{sec:pipeline}

We propose a modular pipeline to systematically evaluate the impact of facial image enhancement on face recognition performance on the propose \textbf{CVBL-CCTV-Face Datset}. As shown in Fig.\ref{fig:pipeline}, the pipeline is organized into three core stages: face detection and alignment, super-resolution enhancement, and embedding extraction and identity matching.

\subsubsection{Face Detection and Alignment}\label{sec:fd_mtcnn} 
The first stage of the pipeline focuses on detecting and geometrically aligning face regions to prepare for subsequent enhancement and recognition. We employ the Multi-task Cascaded Convolutional Neural Network (MTCNN) \cite{mtcnn} for this purpose. MTCNN \cite{mtcnn} operates in three cascaded stages (P-Net, R-Net, O-Net) to progressively refine bounding boxes and localize five facial landmarks for alignment. Figure \ref{fig:mtcnn_challenges} shows six example frames from the CVBL-CCTV-Face Dataset that reveal common failure modes of MTCNN in realistic surveillance scenes. In the top row, the left frame completely misses the face and instead highlights a non facial region, while the middle and right frames produce false positives such as hands and toes. In the bottom row, the left frame again fails to localize the face, and the middle and right continue to misidentify non facial areas. These errors arises from real world factors like low resolution, changing illumination, motion blur, occlusions, and varied poses and illustrate how MTCNN performance degrades outside controlled environments.


\subsubsection{Super-Resolution and Enhancement}
\begin{figure}[!t]
    \centering
    \includegraphics[width=\columnwidth]{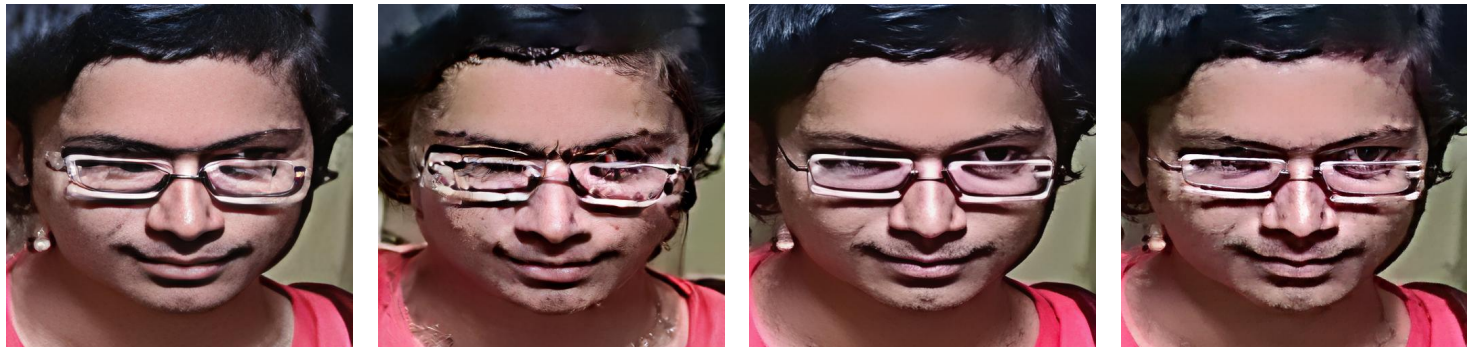}
    \caption{Face enhancement examples using pretrained CodeFormer \cite{codefrmr}, RestoreFormer++ \cite{r++}, PiSA-SR \cite{pisa}, and TSD-SR \cite{tsd} (left to right).}
    \label{fig:enhancement_examples}
\end{figure}

As can be seen from Fig. \ref{fig:pipeline} (part (b)), the aligned LQ face crops are enhanced with a collection of deep learning-based enhancement models specifically trained to address some of the most common surveillance-related image degradations like blur, noise, and compression artifacts. These models comprise \textit{CodeFormer} \cite{codefrmr}, \textit{RestoreFormer++} \cite{r++}, \textit{PiSA-SR} \cite{pisa}, and \textit{TSD-SR} \cite{tsd}. Each of the above-mentioned models addresses some particular problems associated with low-quality face enhancement. \textbf{CodeFormer} \cite{codefrmr} is a codebook-augmented transformer that allows balancing visual sharpness and identity preservation by introducing an easily adjustable fidelity weight. \textbf{RestoreFormer++} \cite{r++} introduces a dynamic key-value memory for high-fidelity prior retrieval with multi-scale feature aggregation. \textbf{PiSA-SR} \cite{pisa} is a dual-LoRA model built upon a pretrained Stable Diffusion backbone, which allows optimizing pixel and semantic fidelity independently. \textbf{TSD-SR} \cite{tsd} aligns super-resolution with face recognition and relies on distillation of target identity scores, making it robust to extreme image degradations. Fig. \ref{fig:enhancement_examples} demonstrates examples of enhancements of LQ faces obtained with different pre-trained models.

    \subsubsection{Embedding Extraction and Recognition} 
    As shown in part (c) of Fig. \ref{fig:pipeline}, super-resolved (SR) images are passed through the \textbf{Inception-ResNet-v1} \cite{inception} network to extract 512-dimensional facial embeddings. This hybrid architecture combines multi-scale Inception modules with residual connections for effective feature representation. The embeddings are $\ell_2$-normalized, ensuring scale-invariant comparisons based on angular similarity. Let $M$ and $P$ denote the number of SR and high-quality (HR) faces, respectively, forming $M \times 512$ and $P \times 512$ matrices. Identity matching is performed by computing Euclidean distances between embeddings: $d(\mathbf{e}_\text{SR}, \mathbf{e}_\text{HR}) = \| \mathbf{e}_\text{SR} - \mathbf{e}_\text{HR} \|_2,
    \label{eq:distance}$. Each SR face is matched to the nearest HR identity, and accuracy is the proportion of correct matches.


\section{Experimental Settings}
\label{sec:experimental_settings}

\subsection{Training Configurations for Face Enhancement Models}

Evaluation of the efficiency of different training approaches to face enhancement and recognition is done using experiments under three setups, each representing a certain degree of domain adaptation. Experiments are performed on the \textbf{CVBL-CCTV-Face Dataset}, as detailed in Section \ref{sec:dataset}. Evaluation procedure is divided into two parts – face enhancement and identity recognition. Face detection and face alignment are performed using MTCNN face detector, while features are extracted using the \textbf{Inception-ResNet-v1} network, 
hybrid architecture pretrained on \textbf{VGGFace2} \cite{vggface2}. 512-dimensional embeddings generated by the network are then used for identity matching. The performance of face recognition is assessed in terms of Top-1 identification accuracy. Experimental setup is as follows:


\subsubsection{Setting 1: Pretrained Models}

The enhancement models are applied to the low-quality (LQ) images in this setting without any fine-tuning. The pretrained weights are being utilized by the models to assess generalization on diverse surveillance conditions. As shown in the first column of Fig. \ref{fig:enhancement_strategies}, pretrained models demonstrate improvements over raw LQ inputs but they often struggle to capture domain-specific characteristics, reducing their robustness in real-world surveillance scenarios.



\subsubsection{Setting 2: Fine-Tuning on Synthetically Degraded HQ Images}

\begin{figure}[!t]
\centering
\includegraphics[width=0.48\textwidth]{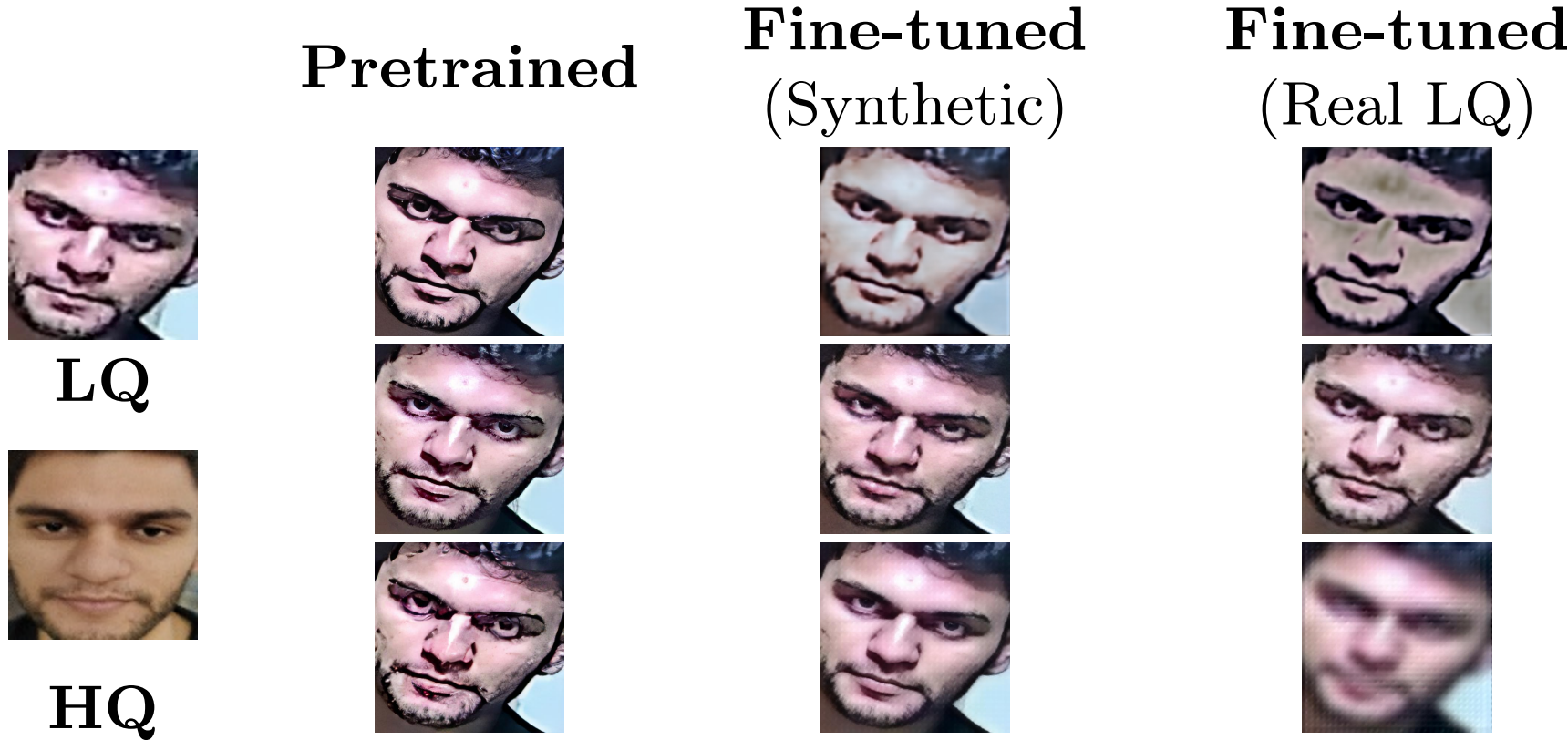}
\caption{Visual comparison between LQ/HQ inputs and restored outputs under different training strategies. Each column shows outputs from CodeFormer, TSD-SR, and RestoreFormer++ models for Pretrained, Fine-tuned on synthetic degradation, and Fine-tuned on real LQ images.}
\label{fig:enhancement_strategies}
\end{figure}

In this configuration, high-quality images from the \textbf{CVBL-CCTV-Face Dataset} are artificially degraded using techniques like Gaussian blur, noise addition, and compression artifacts, which emulate real-world surveillance scenarios. Enhancement networks are fine-tuned using these degraded images so that the network learns common degradations. As can be seen in the second column of Fig. \ref{fig:enhancement_strategies}, the use of artificial degradations for fine-tuning results in substantial improvements in recognition accuracy.

\begin{figure}[!t]
    \centering
    \includegraphics[width=\linewidth]{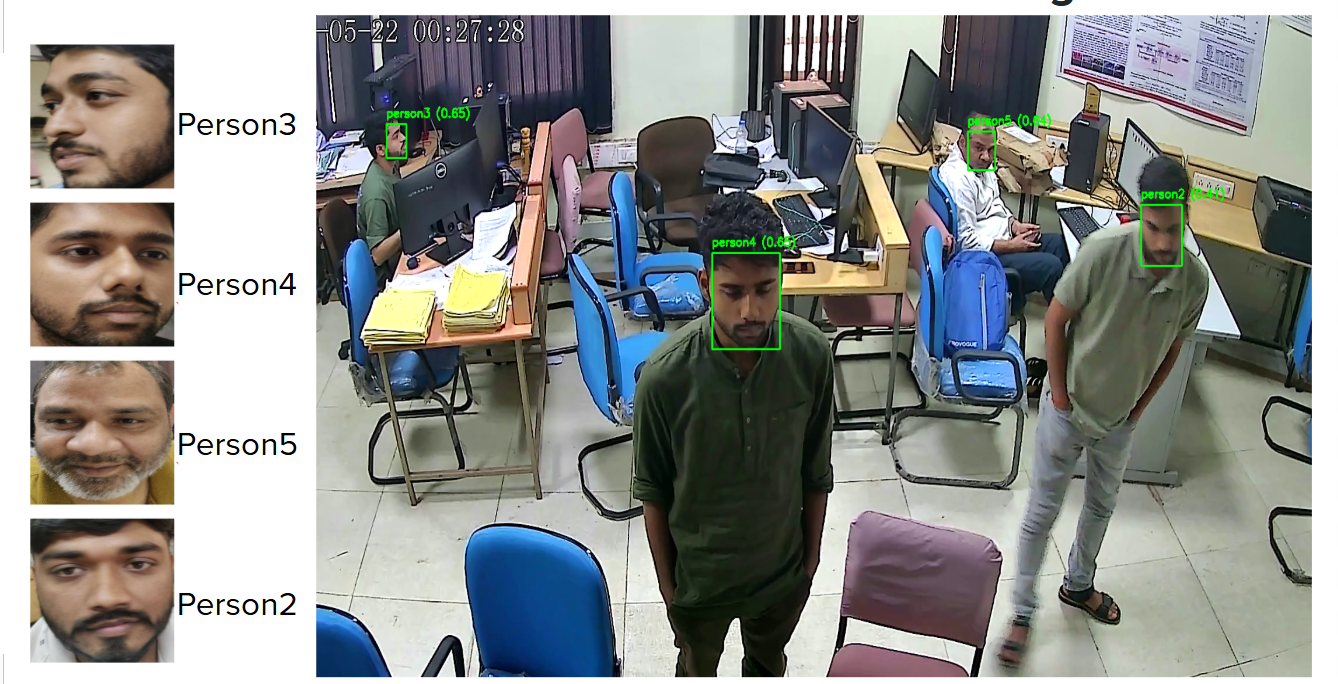}
    \caption{Real-time results: Left shows enrolled HR faces; right shows enhanced live detections with matched identities.}
    
    \label{fig:realtime_face_recognition}
\end{figure}

\subsubsection{Setting 3: Fine-Tuning on Real LQ Images}

In this setting, models are fine-tuned directly on real low-quality surveillance images containing complex artifacts like occlusions, pose variations, motion blur, and lighting changes. As shown in the third column of Fig. \ref{fig:enhancement_strategies}, direct real LQ fine-tuning did not improve over the baseline, indicating that current architectures struggle with uncontrolled real-world distortions, which provides an open challenge motivating future work.

\subsection{Real-Time Enhancement and Recognition System}
\label{sec:realtime_pipeline}

We evaluate the real-time system on the \textbf{CVBL-CCTV-Face Dataset} under diverse live conditions, with enhancement applied as a preprocessing step prior to identity matching. As shown in Fig. \ref{fig:realtime_face_recognition}, a gallery of high-resolution (HR) reference faces is created using Inception-ResNet-v1, producing 512-d embeddings. The faces are detected via MTCNN, enhanced using the proposed models, and passed through Inception-ResNet-v1 to extract embeddings. These are compared to the HR gallery using cosine similarity. The closest match is displayed with a bounding box and identity label.

\begin{figure*}[!t]
    \centering
    \includegraphics[width=\textwidth]{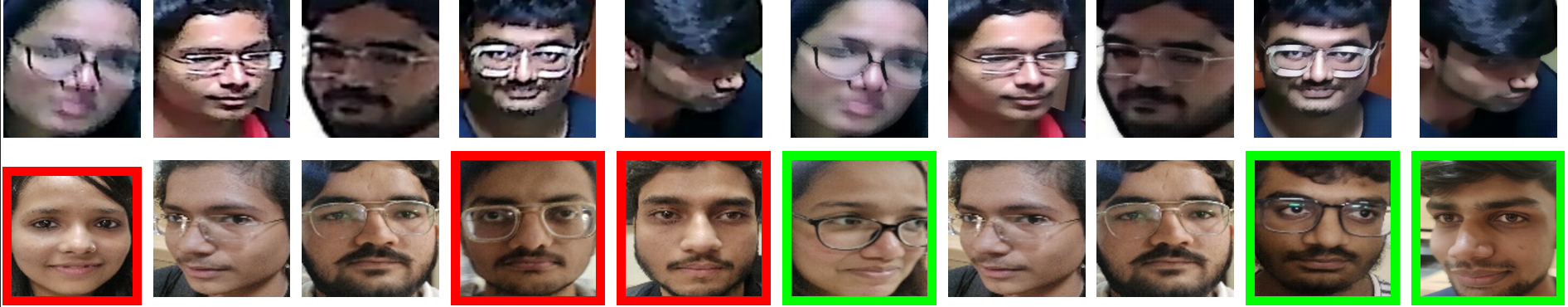} 
    \caption{Facial enhancement and identity recognition results from the \textbf{ CVBL-CCTV-Face Dataset}. Incorrect identity matches are highlighted with red boxes and their corrected matches after enhancement are shown with green boxes.}
    
    \label{fig:face_restoration_results}
\end{figure*}

\section{Results}
\label{sec : results}

\subsection{Face Enhancement and Recognition Performance}
\label{sec:enhancement_performance}

\begin{figure}[!t]
    \centering
    \includegraphics[width=\columnwidth]{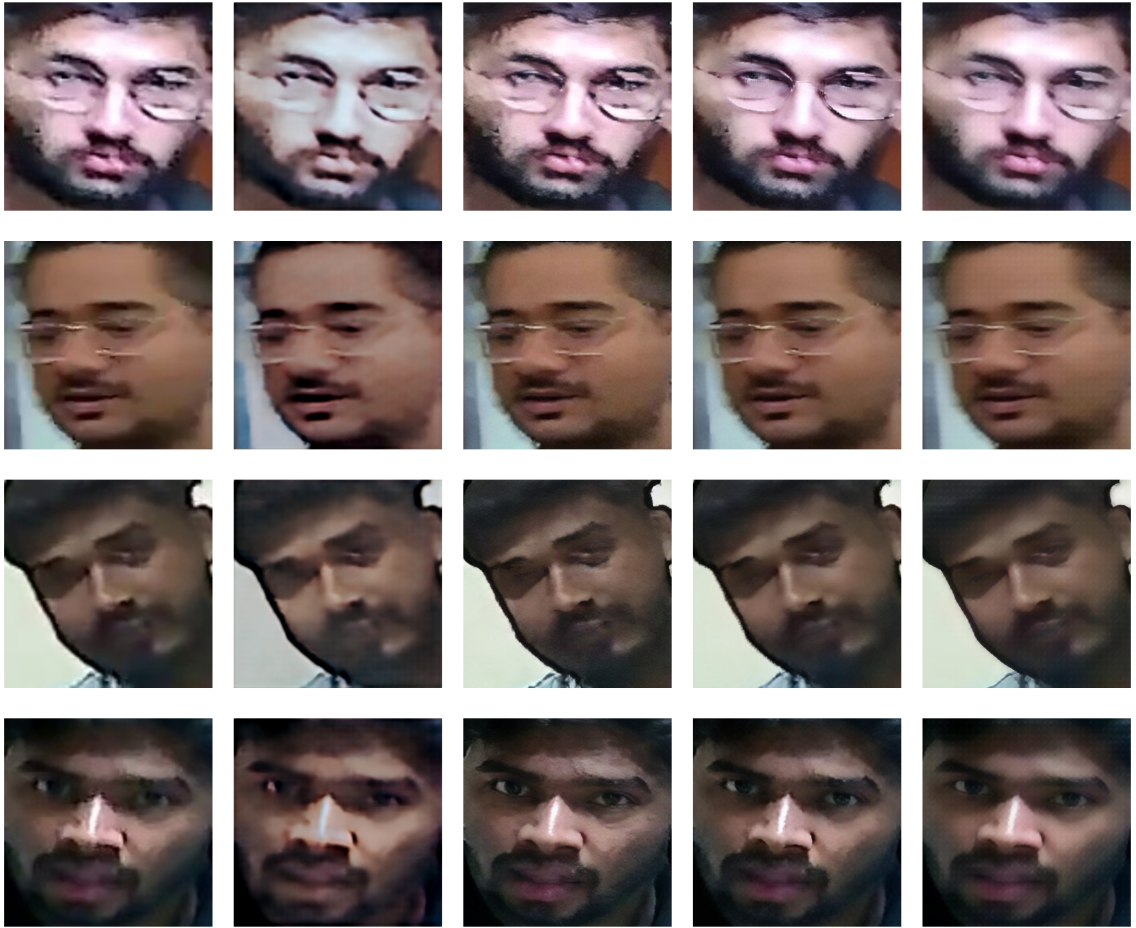}
    \caption{Qualitative face enhancement results on four identities from the \textbf{CVBL-Face dataset}. Each row corresponds to a different individual. From left to right: (1) original low-quality (LQ) surveillance face crop, followed by enhanced outputs from (2) CodeFormer \cite{codefrmr}, (3) RestoreFormer++ \cite{r++}, (4) PiSA-SR \cite{pisa}, and (5) TSD-SR \cite{tsd}.}
    \label{fig:model_comparison}
\end{figure}

As shown in Fig. \ref{fig:face_restoration_results}, the first five columns depict LQ inputs and their HQ references, with red boxes marking incorrect identity matches. The next five columns show the corresponding TSD-SR enhanced outputs and their corrected matches. In column 1, the LQ image was misclassified as ID 32 (\(L_2 = 0.9332\)), but after enhancement, it correctly matches ID 108 (\(L_2 = 0.7562\)) as seen in Column 6. Column 7 retains ID 101, with distance improving from 0.7603 to 0.6988. Column 8 also keeps ID 115, improving from 0.7200 to 0.4700. Column 9 corrects a mismatch as seen in Column 4 from ID 71 (0.8557) to ID 81 (0.8088). Column 10, despite occlusion, recovers ID 100, improving from ID 9 (1.023) to 0.980. These results confirm the enhancement models  effectiveness in reducing embedding gaps and correcting identity mismatches under real-world CCTV conditions. Fig. \ref{fig:model_comparison} further compares CodeFormer, Restore
Former++, PiSA-SR, and TSD-SR (all fine-tuned on synthetic degradations) across four identities. For identity 1, the \(L_2\) distances are 0.890 (CodeFormer), 0.730 (PiSA-SR), 0.930 (RestoreFormer++), and 0.773 (TSD-SR); for identity 2: 0.680, 0.620, 0.617, and 0.604; identity 3: 0.841, 0.824, 0.766, and 0.795; and identity 4: 0.748, 0.707, 0.668, and 0.679 respectively. Overall, fine-tuning consistently improves identity similarity, with TSD-SR and RestoreFormer++ preserving identities most effectively.

\subsection{Real-Time Face Enhancement and Recognition Results}
\label{sec:realtime_results}
We evaluated on an NVIDIA GeForce RTX 3090 GPU at approximately 25 FPS under diverse CCTV conditions, including motion blur, occlusion, illumination changes, and non-frontal poses. Fig. \ref{fig:realtime_face_recognition} illustrates the results, where the left panel shows high-resolution (HR) enrolled faces and the right panel shows live detections. Detected low-quality (LQ) faces are enhanced on-the-fly, and embeddings are extracted and matched to the HR gallery using dot product similarity. Predicted identities are annotated with similarity scores: \textit{Person3} (0.65), \textit{Person4} (0.66), \textit{Person5} (0.64), and \textit{Person2} (0.41). The incorrect match of \textit{Person2} highlights sensitivity to ambiguous low-quality inputs, yet overall results validate enhancement as a necessary preprocessing step for real-time recognition under surveillance degradations.

\subsection{Evaluation Under Different Training Configurations}
\label{sec:results_discussion}

 We evaluated three enhancement strategies on \textbf{CVBL-CCTV-Face Dataset} with realistic degradations from surveillance, shown in Table \ref{tab:restoration_results}. The pipeline contains face detection (MTCNN), enhancement (Transformer and Diffusion-based models) and feature extraction (Inception-ResNet-v1 trained on VGGFace2). We evaluated the performance by means of $L_2$ distance and top-1 identification accuracy. Baseline (without enhancement) yields 62.6\% identification accuracy.

\subsubsection{Evaluation with Pre-trained Models}
The identification rate for PiSA-SR and TSD-SR is 64.77\% and 64.45\%, respectively, using publicly pre-trained models without any fine-tuning, which is slightly higher than baseline and demonstrates poor generalization on real CCTV degradations.

\subsubsection{Fine Tuning on Real Low Quality Images}
Different fine tuning approaches yield different identification rates, with RestoreFormer++ getting 62.75\%, CodeFormer and TSD-SR obtaining 40.20\% and 33.33\%, respectively. This difference can be explained by the fact that real LQ images are captured independently, are spatially unaligned and have multiple uncontrolled degradations, unlike synthetic degraded HQ images.

\subsubsection{Fine-Tuning on Synthetically Degraded HQ Images}
Synthetic fine tuning proved to be the most successful: TSD-SR obtains 70.08\% accuracy, PiSA-SR - 67.81\% and RestoreFormer++ - 67.49\%.

\begin{table}[!t]
\caption{Benchmark Comparison of Representative Enhancement Methods Using Top-1 Recognition Accuracy (\%)}
\begin{center}
\setlength{\tabcolsep}{3pt} 
\begin{tabular}{|c|c|c|c|}
\hline
\textbf{Model} & \multicolumn{3}{|c|}{\textbf{Training Setting}} \\
\cline{2-4} 
\textbf{} & \textbf{\textit{Pre-Trained}} & \textbf{\textit{FT (Real LQ)}} & \textbf{\textit{FT (Synthetic)}} \\
\hline
PiSA-SR\cite{pisa}        & 64.77 & 3.23  & 67.81 \\
\hline
TSD-SR\cite{tsd}          & 64.45 & 33.33 & 70.08 \\
\hline
CodeFormer\cite{codefrmr} & 57.69 & 40.20 & 59.20 \\
\hline
RestoreFormer++\cite{r++} & 59.46 & 62.75 & 67.49 \\
\hline
\multicolumn{4}{l}{$^{\mathrm{a}}$Baseline (No enhancement): 62.60\%.}
\end{tabular}
\label{tab:restoration_results}
\end{center}
\end{table}


\section{Conclusion}

The CVBL-CCTV-Face Dataset can be considered as an essential benchmark that can be used to improve face enhancement and recognition in real-world surveillance situations. The pipeline created using this dataset, consisting of face detection, deep learning-based image enhancement, and identity recognition, performs well in different conditions. Of all the training methods tested, those involving synthetic degradations of high-resolution images resulted in the best recognition accuracy, which proves the validity of degradation modeling. Yet, the problem is that the actual scenario in real-world surveillance is associated with low-quality real-life degradations, which are much more complicated and unpredictable. Models cannot cope with this situation at the moment, yet it is a promising way forward. Our real-time system proves the feasibility of the suggested pipeline, and these findings can become the foundation for developing face recognition systems.

\section{Future Work}
We will extend the identification benchmark by adding verification metrics such as FAR/FRR, inference latency, and computational cost across different hardware platforms in our future work.

\section*{Acknowledgment}
This work was supported by the I-DAPT Hub Foundation, IIT (BHU) Varanasi.
We acknowledge the fellowship and research support from Indian Institute of Information Technology Allahabad. We sincerely thank Mr. Hritu Raj, Mr. Vikas Rajput, Mr. Samar Borkar, Mr. Anubhav, Mr. Shirshak Mandal, Mr. Govind Anand, and Mr. Prathmesh Pampattiwar for their support in creation of the dataset used in this study.

\bibliographystyle{ieeetr}
\bibliography{ref}

\begin{thebibliography}{10}

\bibitem{mtcnn}
K.~Zhang, Z.~Zhang, Z.~Li, and Y.~Qiao, ``Joint face detection and alignment using multitask cascaded convolutional networks,'' {\em IEEE signal processing letters}, vol.~23, no.~10, pp.~1499--1503, 2016.

\bibitem{5}
C.~Dong, C.~C. Loy, K.~He, and X.~Tang, ``Learning a deep convolutional network for image super-resolution,'' in {\em Computer Vision--ECCV 2014: 13th European Conference, Zurich, Switzerland, September 6-12, 2014, Proceedings, Part IV 13}, pp.~184--199, Springer, 2014.

\bibitem{10}
M.~Haris, G.~Shakhnarovich, and N.~Ukita, ``Deep back-projection networks for super-resolution,'' in {\em Proceedings of the IEEE conference on computer vision and pattern recognition}, pp.~1664--1673, 2018.

\bibitem{22}
J.~Kim, J.~K. Lee, and K.~M. Lee, ``Deeply-recursive convolutional network for image super-resolution,'' in {\em Proceedings of the IEEE conference on computer vision and pattern recognition}, pp.~1637--1645, 2016.

\bibitem{27}
J.~Liang, J.~Cao, G.~Sun, K.~Zhang, L.~Van~Gool, and R.~Timofte, ``Swinir: Image restoration using swin transformer,'' in {\em Proceedings of the IEEE/CVF international conference on computer vision}, pp.~1833--1844, 2021.

\bibitem{52}
Z.~Wang, A.~C. Bovik, H.~R. Sheikh, and E.~P. Simoncelli, ``Image quality assessment: from error visibility to structural similarity,'' {\em IEEE transactions on image processing}, vol.~13, no.~4, pp.~600--612, 2004.

\bibitem{65}
K.~Zhang, J.~Liang, L.~Van~Gool, and R.~Timofte, ``Designing a practical degradation model for deep blind image super-resolution,'' in {\em Proceedings of the IEEE/CVF international conference on computer vision}, pp.~4791--4800, 2021.

\bibitem{50}
X.~Wang, L.~Xie, C.~Dong, and Y.~Shan, ``Real-esrgan: Training real-world blind super-resolution with pure synthetic data,'' in {\em Proceedings of the IEEE/CVF international conference on computer vision}, pp.~1905--1914, 2021.

\bibitem{17}
J.~Johnson, A.~Alahi, and L.~Fei-Fei, ``Perceptual losses for real-time style transfer and super-resolution,'' in {\em Computer Vision--ECCV 2016: 14th European Conference, Amsterdam, The Netherlands, October 11-14, 2016, Proceedings, Part II 14}, pp.~694--711, Springer, 2016.

\bibitem{24}
C.~Ledig, L.~Theis, F.~Husz{\'a}r, J.~Caballero, A.~Cunningham, A.~Acosta, A.~Aitken, A.~Tejani, J.~Totz, Z.~Wang, {\em et~al.}, ``Photo-realistic single image super-resolution using a generative adversarial network,'' in {\em Proceedings of the IEEE conference on computer vision and pattern recognition}, pp.~4681--4690, 2017.

\bibitem{48}
X.~Wang, K.~Yu, S.~Wu, J.~Gu, Y.~Liu, C.~Dong, Y.~Qiao, and C.~Change~Loy, ``Esrgan: Enhanced super-resolution generative adversarial networks,'' in {\em Proceedings of the European conference on computer vision (ECCV) workshops}, pp.~0--0, 2018.

\bibitem{25}
W.~Li, K.~Zhou, L.~Qi, L.~Lu, and J.~Lu, ``Best-buddy gans for highly detailed image super-resolution,'' in {\em Proceedings of the AAAI Conference on Artificial Intelligence}, vol.~36, pp.~1412--1420, 2022.

\bibitem{70}
Y.~Zhang, B.~Ji, J.~Hao, and A.~Yao, ``Perception-distortion balanced admm optimization for single-image super-resolution,'' in {\em European Conference on Computer Vision}, pp.~108--125, Springer, 2022.

\bibitem{43}
L.~Sun, J.~Liang, S.~Liu, H.~Yong, and L.~Zhang, ``Perception-distortion balanced super-resolution: A multi-objective optimization perspective,'' {\em IEEE Transactions on Image Processing}, 2024.

\bibitem{49}
X.~Wang, K.~Yu, C.~Dong, X.~Tang, and C.~C. Loy, ``Deep network interpolation for continuous imagery effect transition,'' in {\em Proceedings of the IEEE/CVF Conference on Computer Vision and Pattern Recognition}, pp.~1692--1701, 2019.

\bibitem{29}
J.~Liang, H.~Zeng, and L.~Zhang, ``Details or artifacts: A locally discriminative learning approach to realistic image super-resolution,'' in {\em Proceedings of the IEEE/CVF conference on computer vision and pattern recognition}, pp.~5657--5666, 2022.

\bibitem{46}
J.~Wang, Z.~Yue, S.~Zhou, K.~C. Chan, and C.~C. Loy, ``Exploiting diffusion prior for real-world image super-resolution,'' {\em International Journal of Computer Vision}, vol.~132, no.~12, pp.~5929--5949, 2024.

\bibitem{30}
X.~Lin, J.~He, Z.~Chen, Z.~Lyu, B.~Dai, F.~Yu, Y.~Qiao, W.~Ouyang, and C.~Dong, ``Diffbir: Toward blind image restoration with generative diffusion prior,'' in {\em European Conference on Computer Vision}, pp.~430--448, Springer, 2024.

\bibitem{pisa}
L.~Sun, R.~Wu, Z.~Ma, S.~Liu, Q.~Yi, and L.~Zhang, ``Pixel-level and semantic-level adjustable super-resolution: A dual-lora approach,'' {\em arXiv preprint arXiv:2412.03017}, 2024.

\bibitem{tsd}
L.~Dong, Q.~Fan, Y.~Guo, Z.~Wang, Q.~Zhang, J.~Chen, Y.~Luo, and C.~Zou, ``Tsd-sr: One-step diffusion with target score distillation for real-world image super-resolution,'' {\em arXiv preprint arXiv:2411.18263}, 2024.

\bibitem{51}
Y.~Wang, W.~Yang, X.~Chen, Y.~Wang, L.~Guo, L.-P. Chau, Z.~Liu, Y.~Qiao, A.~C. Kot, and B.~Wen, ``Sinsr: diffusion-based image super-resolution in a single step,'' in {\em Proceedings of the IEEE/CVF conference on computer vision and patZhangtern recognition}, pp.~25796--25805, 2024.

\bibitem{56}
R.~Wu, L.~Sun, Z.~Ma, and L.~Zhang, ``One-step effective diffusion network for real-world image super-resolution,'' {\em Advances in Neural Information Processing Systems}, vol.~37, pp.~92529--92553, 2024.

\bibitem{gfpgan}
X.~Wang, Y.~Li, H.~Zhang, and Y.~Shan, ``Towards real-world blind face restoration with generative facial prior,'' in {\em Proceedings of the IEEE/CVF conference on computer vision and pattern recognition}, pp.~9168--9178, 2021.

\bibitem{vqfr}
Y.~Gu, X.~Wang, L.~Xie, C.~Dong, G.~Li, Y.~Shan, and M.-M. Cheng, ``Vqfr: Blind face restoration with vector-quantized dictionary and parallel decoder,'' in {\em European Conference on Computer Vision}, pp.~126--143, Springer, 2022.

\bibitem{codefrmr}
S.~Zhou, K.~Chan, C.~Li, and C.~C. Loy, ``Towards robust blind face restoration with codebook lookup transformer,'' {\em Advances in Neural Information Processing Systems}, vol.~35, pp.~30599--30611, 2022.

\bibitem{r++}
Z.~Wang, J.~Zhang, T.~Chen, W.~Wang, and P.~Luo, ``Restoreformer++: Towards real-world blind face restoration from undegraded key-value pairs,'' {\em IEEE Transactions on Pattern Analysis and Machine Intelligence}, vol.~45, no.~12, pp.~15462--15476, 2023.

\bibitem{inception}
C.~Szegedy, S.~Ioffe, V.~Vanhoucke, and A.~Alemi, ``Inception-v4, inception-resnet and the impact of residual connections on learning,'' in {\em Proceedings of the AAAI conference on artificial intelligence}, vol.~31, 2017.

\bibitem{vggface2}
Q.~Cao, L.~Shen, W.~Xie, O.~M. Parkhi, and A.~Zisserman, ``Vggface2: A dataset for recognising faces across pose and age,'' in {\em 2018 13th IEEE international conference on automatic face \& gesture recognition (FG 2018)}, pp.~67--74, IEEE, 2018.

\end{thebibliography}

\end{document}